\documentclass[lettersize,journal]{IEEEtran}
\IEEEoverridecommandlockouts
\usepackage{cite}
\usepackage{amsmath,amssymb,amsfonts}
\usepackage{algorithmic}
\usepackage{graphicx}
\usepackage{textcomp,subcaption}
\usepackage{xcolor,multicol,multirow}
\usepackage{enumerate,textcomp,array,enumitem}
\usepackage{placeins}
\usepackage{threeparttable,url}
\usepackage{dblfloatfix}
\usepackage{flushend}
\usepackage{setspace}
\usepackage{soul}
\usepackage{algorithm}
\usepackage[running,displaymath]{lineno}

\def\BibTeX{{\rm B\kern-.05em{\sc i\kern-.025em b}\kern-.08em
    T\kern-.1667em\lower.7ex\hbox{E}\kern-.125emX}}

\begin{document}

\title{Semi-Supervised Multi-Task Learning Based Framework for Power System Security Assessment}
  
\author{
    \IEEEauthorblockN{M. E. Za'ter\IEEEauthorrefmark{1}, A. Sajadi\IEEEauthorrefmark{1}\IEEEauthorrefmark{2}, B. M. Hodge\IEEEauthorrefmark{1}\IEEEauthorrefmark{2}}\\
    \IEEEauthorblockA{\IEEEauthorrefmark{1}University of Colorado Boulder, Boulder, CO 80303}\\
    \IEEEauthorblockA{\IEEEauthorrefmark{2}National Renewable Energy Laboratory (NREL), Golden, CO 80401}
}

%\thanks{Muhy Eddin Za\'ter is with the University of Colorado Boulder, USA, Corresponding Author: muhy.zater@colorado.edu}

%\thanks{A.~Sajadi and B.~M.~Hodge are with the  University of Colorado Boulder, Boulder, CO 80309 USA, and the  National
%Renewable Energy Laboratory (NREL), Golden, CO 80401 USA (e-mail:
%Amir.Sajadi@\{colorado.edu,nrel.gov\},  BriMathias.Hodge@colorado.edu and Bri.Mathias.Hodge@nrel.gov}

%\thanks{M.~E.~Za'ter is with the Department of Electrical, Computer and Energy Engineering at the University of Colorado Boulder, 425 UCB, Boulder, CO 80309, USA, Email: muhy.zater@colorado.edu}

%\thanks{A.~Sajadi is with the Renewable and Sustainable Energy Institute (RASEI) at the University of Colorado Boulder, 4001 Discovery Drive, Boulder, CO 80303, USA, Email: Amir.Sajadi@colorado.edu}

%\thanks{B.~M.~Hodge is with the Department of Electrical, Computer and Energy Engineering, the Department of Applied Mathematics, and the Renewable and Sustainable Energy Institute (RASEI) at the University of Colorado Boulder, 425 UCB, Boulder, CO 80309, USA, and the Grid Planning and Analysis Center at the National Renewable Energy Laboratory (NREL), 15013 Denver W Pkwy, Golden, CO 80401, USA, Email: BriMathias.Hodge@colorado.edu, Bri.Mathias.Hodge@nrel.gov}

% make the title area
\maketitle
%\begin{abstract} %% This section will be written the last
%The rapid integration of renewable energy sources and inverter-based resources within the electrical grid is challenging the foundation of power system security. 
%This paper develops a novel machine learning-based framework for power system dynamic security assessment that is accurate, reliable, and topology-aware, which provides a confidence measure for its predictions. The kernel algorithm behind the proposed framework leverages conditional encoders and multi-task learning. Computer simulations of the IEEE 68-bus system validate our method, utilizing two distinct database generation techniques. The results indicate the superior performance of our algorithm when compared with the existing state-of-the-art machine learning methodologies for security assessment in terms of accuracy, robustness and the ability to scale on larger power systems. Our work highlights the value of employing auto-encoders for security assessment in terms of accuracy, reliability, and robustness. All datasets and codes used in this paper have been made available to the public to make the results of the paper reproducible and fully transparent. 
%\end{abstract}
\begin{abstract}
    This paper develops a novel machine learning-based framework using Semi-Supervised Multi-Task Learning (SS-MTL) for power system dynamic security assessment that is accurate, reliable, and aware of topological changes. The learning algorithm underlying the proposed framework integrates conditional masked encoders and employs multi-task learning for classification-aware feature representation, which improves the accuracy and scalability to larger systems. Additionally, this framework incorporates a confidence measure for its predictions, enhancing its reliability and interpretability. A topological similarity index has also been incorporated to add topological awareness to the framework. Various experiments on the IEEE 68-bus system were conducted to validate the proposed method, employing two distinct database generation techniques to generate the required data to train the machine learning algorithm. The results demonstrate that our algorithm outperforms existing state-of-the-art machine learning based techniques for security assessment in terms of accuracy and robustness. Finally, our work underscores the value of employing auto-encoders for security assessment, highlighting improvements in accuracy, reliability, and robustness. All datasets and codes used have been made publicly available to ensure reproducibility and transparency.
\end{abstract}
\begin{IEEEkeywords}
machine learning, multi-tasking learning, power system stability, power system security assessment
\end{IEEEkeywords}

\section{Introduction}

\IEEEPARstart{P}{ower} system security assessment requires high accuracy, robustness, scalability, and interpretability \cite{alimi2020review}, and the lack thereof could result in interruptions of power delivery and potentially lead to disruption of service, or blackouts. Recently, with the consistent increase in the periods of operation dominated by variable renewable energy, it has become essential to enhance the security assessment of the power system to accommodate higher degrees of uncertainty, variability, and complexity, which are mainly introduced by the nature of solar and wind generation \cite{panciatici2010security, sajadi2020identification}. Such security assessment system improvements are pivotal to assist system operators in determining near real-time corrective actions to maintain system security throughout the year \cite{CREMER2021106571}.

Solutions for dynamic security assessment have evolved in two main directions; model-based and model-free. Model-based techniques rely on mathematical models of the power system to assess its security. They use models of power system components to derive explicit nonlinear differential equations that describe a power system, including its different components, and solve these equations in closed form. Although they provide relatively accurate solutions depending on the granularity of the model used, they have huge computational burdens, thus offering limited applicability for large power systems \cite{chiang2011direct}. Similarly, electromagnetic transient (EMT) models are considered an accurate option for dynamic security assessment, but they are also only practical for small-scale system, due to their high computational complexity\cite{kenyon2023comparison}. As an alternative to model-based techniques model-free techniques have been developed which do not rely on explicit mathematical models of the power system \cite{fan2018post}. These algorithms learn directly from historical and/or simulated data to assess the security of the power system. In the model-free approach, the system is analyzed by observing the patterns and relationships in the data, without requiring an explicit understanding of the underlying system. These techniques are typically faster and computationally cheaper, and hence scalable to larger systems, but are often not as accurate as model-based techniques as they are approximations of the actual mathematical representations \cite{duchesne2018using}. Nonetheless, they are practical for real-time operational purposes due to their computational efficiency, and therefore their potential has fueled researchers efforts into developing more accurate and reliable model-free methods \cite{gurusinghe2015post, deng2011real}.

Over the past decade, model-free techniques for power system security assessment have converged chiefly towards the use of machine learning (ML) \cite{arteaga2019deep, gholami2020static, sevilla2015static}, mainly encouraged by the wider adoption of advanced measurement instruments, e.g., especially Phasor Measurement Units (PMUs) \cite{zhang2019deep}. This sensing technology has enabled an unprecedented degree of visibility into the power grid by generating an abundance of data, presenting immense opportunities for data-driven methodologies within power system applications \cite{wang2020machine}. A variety of ML-based techniques, such as decision trees (DT) and random forests (RF), have been investigated for power system security assessment \cite{genc2010decision, wehenkel1993decision, sunitha2013online, vasconcelos2016online, liu2013systematic}. Although these methods offer high interpretability, which is highly desirable for power system operators, they tend to fall short in terms of achieving the desired accuracy, robustness, and scalability \cite{lecun2015deep, aguero2017modernizing}. Therefore, alternative solutions such as support vector machines (SVM)\cite{gomez2010support}, logistic regression \cite{zhang2019deep}, long short-term memory \cite{copp2017time}, and ensemble methods \cite{liu2018accurate, gomez2010support} have been explored to attain improved accuracy. Nevertheless, when these algorithms are applied to larger power systems, such as those typically encountered in real-world scenarios, their accuracy fails to meet the expected performance standards \cite{zhang2019deep, genc2010decision, konstantelos2018using, copp2017time}.

For applications such as power system security assessment that are safety-critical and continuously changing (i.e different contingencies and topologies), it is crucial to have high accuracy, stability against minor disturbances and adversarial threats, awareness of system topology, and the capacity to handle a broad range of possible scenarios from type of contingency to different topologies. ML techniques for power system security assessment strive to achieve a balance between interpretability and scalability to effectively support the ever-evolving needs of the power industry \cite{xiao2006contingency}. Interpretability is important to trace the decision-making process \cite{venzke2021efficient} and recommend corrective actions,while scalability is important because power systems are continuously subject to topological changes, arising from various factors such as maintenance activities or environmental conditions \cite{zhang2021confidence}. 
%
%Given the above-mentioned requirements, and in addition to the surge of data, and the advances made in other domains using advanced machine learning and deep learning techniques.
With all of these demanding requirements for power system security assessment, conventional machine learning models have not been able to satisfy the desired performance requirements \cite{ahmadi2012maximum, diaz2013static}. However, with the huge surge of data and digitized measurements in power system, it became feasible to apply more advanced, data-hungry algorithms of machine learning and deep learning that have proven to be very effective in other fields.
These advanced methods, including the application of deep auto-encoders \cite{zhang2019deep}, Bayesian methods \cite{zhang2021confidence}, and tree-regularization techniques \cite{ren2021interpretable}, have been utilized to push the boundaries of what is achievable in power system security assessment for larger systems.
ML methods thus far have demonstrated promising utility to tackle power system security assessment, exhibiting the capacity to accommodate numerous contingencies and scenarios in offline computations, surpassing the capabilities of deterministic and probabilistic approaches in terms of the number of scenarios and contingencies considered \cite{mishra2012contingency}. However, despite their high accuracy, these models have not yet fully satisfied the stringent critical requirements inherent to the security assessment of power systems \cite{ren2021vulnerability}.

In this paper we develop a new framework for dynamic security assessment in power systems. At the core of this framework resides an ML-based technique that leverages power-system-specific features for enhanced awareness and uses a deep autoencoder for feature extraction. The proposed algorithm's training technique employs multi-task learning and semi-supervised learning approaches to develop a classification-aware conditional deep autoencoder, customizing the feature representation to enhance the classifier. The results demonstrate the proposed framework outperforming the existing state-of-the-art methodologies, achieving a 5\% improvement in F1-score over the method in \cite{zhang2021confidence}, and a 7.2\% increase in F1-score compared to \cite{arteaga2019deep} when evaluated on the IEEE 68-bus system. This performance was also maintained when tests included bad data. All generated datasets and the code for the proposed algorithms have been made available to the public at no cost for the sake of reproducibility \cite{githubGitHubMuhizatarMachinelearningforpowersystemsecurityassessment}.

The contributions of this paper are three-fold. \textit{First}, we present an enhanced classifier that outperforms traditional methods like the Fisher score \cite{fliscounakis2013contingency}, Principal Component Analysis (PCA) \cite{liu2018accurate}, and deep learning-based classifiers \cite{zhang2021confidence, meegahapola2021power}, in addition to integrating a conditional deep autoencoder. This integration enhances the model's contingency awareness, improving power system security assessment and leading to better accuracy and sensitivity \cite{zhang2019deep}. \textit{Second}, we shift from the Bayesian inference proposed in \cite{zhang2021confidence, liu2020bayesian} to a feed-forward neural network (FFNN) classifier that embeds a novel distance measure based on the Mahalanobis distance \cite{fan2018post, mili1993robust} specifically tailored for power systems. This twofold enhancement not only sidesteps the noise robustness and scalability challenges inherent in Bayesian inference \cite{zhang2021confidence}, but also ensures a more reliable and robust classifier that deeply considers power system characteristics. \textit{Finally}, we address a substantial gap in the current literature by introducing a method that amplifies topology awareness grounded in power system attributes, offering a more discerning measure of topological differences compared to prevalent empirical methods \cite{zhang2021confidence} using Singular Value Sequence (SVS), a method known for effectively capturing the characteristics of power systems \cite{fan2018post}.

%The remainder of this paper is organized as follows. Section II lays out a brief introduction about the problem. In Section III, we introduced the developed machine learning framework. While Section IV presents the implementation of the different components of the framework. Finally section V shows the results of the experiments on the developed framework compared with other existing machine learning based algorithms.
%\input{New_sections/2_problem_formulation}
%\input{New_sections/2_Literature}
\section{Proposed Framework and Associated Processes and Algorithms}

\begin{figure}[h]
    \centering
    \includegraphics[width=.49\textwidth, height=4.5cm]{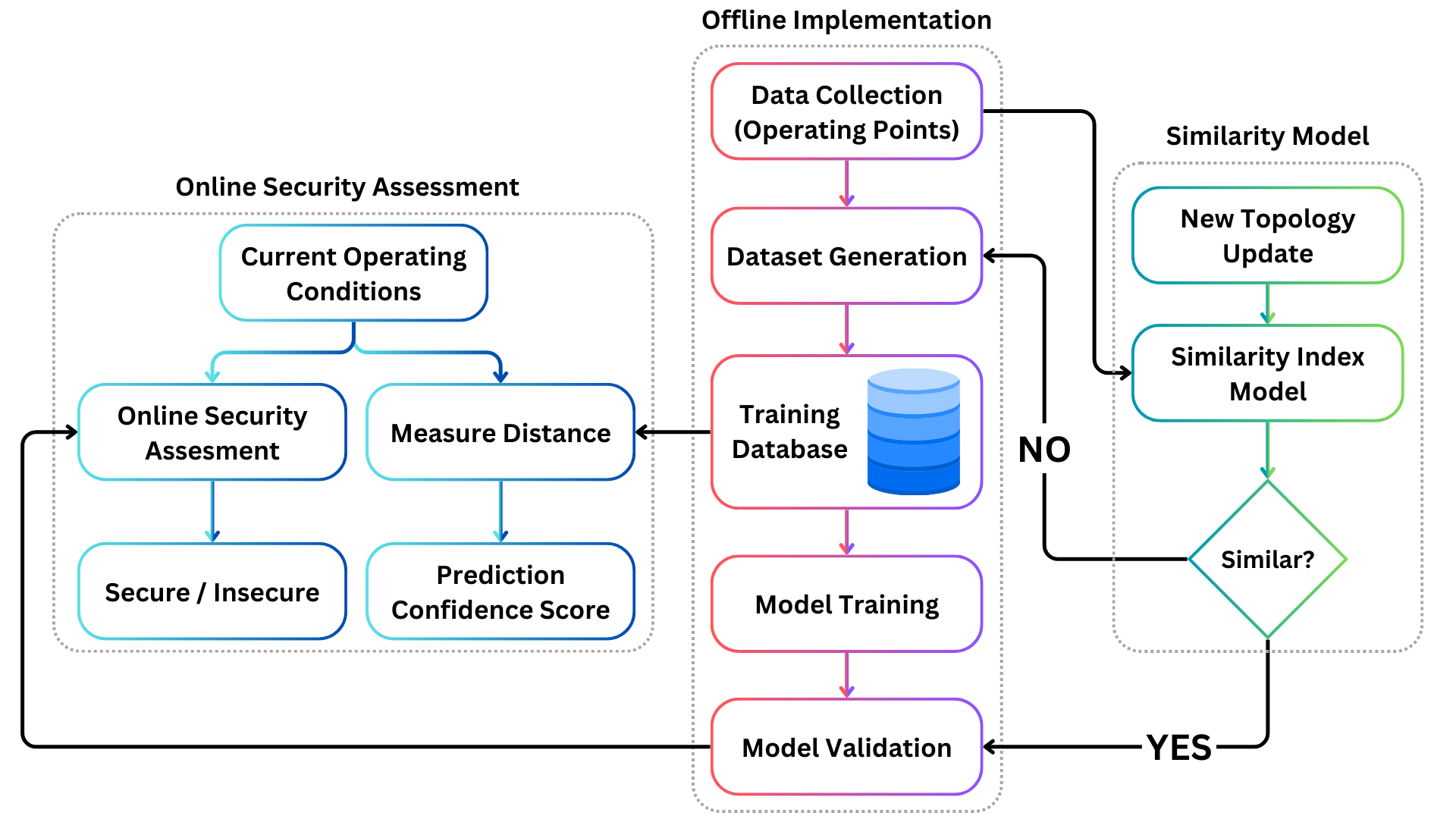}
    \caption{The proposed Machine Learning framework for Dynamic Security Assessment (DSA)}
    \label{fig:proposed_method}
\end{figure}

In this section, we discuss the structure and processes of our novel ML-based framework for dynamic security assessment. Our framework approaches this complex nonlinear problem in a multi-faceted manner, addressing the challenges in database generation, limitations of machine learning algorithms, in addition to the complex and nonlinear nature of power systems, as illustrated in Fig. \ref{fig:proposed_method}. 

As illustrated in \ref{fig:proposed_method}, we divide the proposed framework into two main stages; offline implementation, and online security assessment. The offline stage is conducted before using the model for classifying the state of the system and consists of preparing and generating the training dataset, in addition to the model training and evaluation. The online stage includes classifying the current and new operating conditions into secure and insecure conditions.

The offline stage starts with data collection on diverse system characteristics; a complex database is created using computer simulation to manage the data collected from power system dynamic simulations, which are crucial for training the machine learning algorithms (\ref{generation}). After the dataset is generated, the machine learning models are trained on these datasets (\ref{ML}), which are then evaluated across different operating conditions on a testset to ensure that the trained algorithm performs as desired \ref{evaluation}). Whenever a new power system topology is introduced the similarity model component assesses whether the topological changes necessitate database updates and re-training the machine learning model (\ref{similarity-sec}). This step helps in improving the topological awareness of the model, especially improving its ability to minimize missed alarms, as described in Section \ref{similarity-sec}.

In the online stage, the trained model is used to classify new operating conditions into secure or insecure states. With every state prediction generated from the model a distance measure component provides confidence levels for the predictions to provide a level of interpretability for operators (\ref{distance}). The remainder of this section explore each of the components mentioned above in detail.

\subsection{Data Collection and Assumptions}

Data collection is a critical phase in any machine learning-based algorithm process, as it is essential for training and testing. The more comprehensive, well-distributed, and representative of the various scenarios that the model might encounter the training data is, the more reliable, robust, and accurate the performance of the ML algorithms will be.

Here, we rely on commercially available power system simulation tools (ANDES \cite{cui2018andes}) to generate our data. The simulations are tailored to cover a vast range of system operating conditions and scenarios, ensuring that the dataset is diverse and comprehensive as further discussed in \ref{generation}. We consider this data in lieu of high-precision high-frequency measurement data that are used in real-time operations, such as Phasor Measurement Units (PMUs). Therefore, parameters such as voltage levels, phase shifts, and real and reactive power are all computed and integrated into the dataset.

%As a test case, we utilize the IEEE 68-bus system model as referenced in \cite{pal2006robust}. This system serves as a benchmark to validate the effectiveness of the proposed framework. finally, to diversify the operating conditions, we simulate additional scenarios within a predetermined distribution range, ensuring a robust evaluation of the system under various conditions.

\subsection{Database Generation}
\label{generation}
Ideally, machine learning models are trained on historical data of the power systems, as historical data offers exact information on previous operating conditions. However, historical data does not always cover a wide range of the potential operating points, and often contains few insecure points, which are insufficient to train an accurate data-driven technique. Therefore, constructing a database beyond the available historical data is crucial to build a machine learning based system that is accurate and reliable for system security assessment. In this work, we employ two database construction methodologies as detailed below.

\subsubsection{Time-Domain-Simulation}

\begin{figure}[h]
\centering
\includegraphics[width=0.45\textwidth]{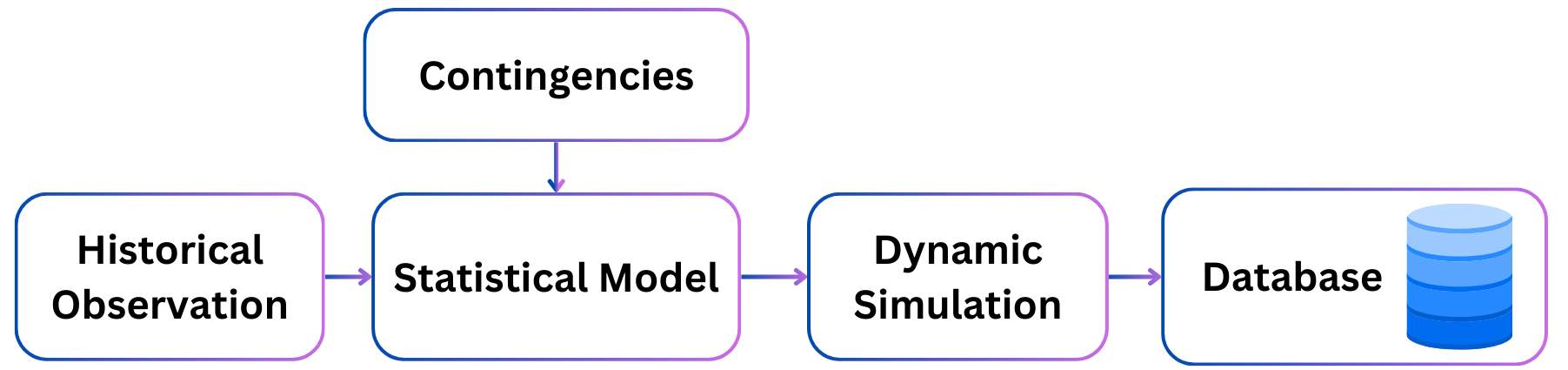}
\caption{Process flow diagram of time domain simulations to create a database}
\label{fig:flow-TDS}
\end{figure}

Time-domain-Simulation (TDS) is the most common method for database generation in the literature \cite{copp2017time} as it accurately determines both secure and insecure operating points, and can take into consideration many aspects of security such as transient, voltage, static, and small-signal stability.
The process of generating the database using TDS largely depends on the nature and availability of historical data and observations, as well as user choices. However, the majority of the literature \cite{sevilla2015static, alimi2020review, alimi2019real} adheres to the flow shown in Fig \ref{fig:flow-TDS} which what is followed in this work.

Initially, power system configuration data and available historical observations are collected. Subsequently, a statistical model is applied to these historical observations to generate a larger number of samples to cover more (and more extreme) potential operating points \cite{tomin2016machine}. These samples are then considered as pre-fault operating conditions used to perform time-domain simulations after setting up specific contingencies of interest \cite{glover2012power}. In this work, we attempt to consider more than one aspect of power system and not to be limited to one, such as transient or voltage security \cite{zhou2021transient, gholami2020static, hagmar2020voltage}. The labeling of secure and insecure for each operating state is based on the following four criteria:

\begin{itemize}
\item \textbf{Transient Stability:} For each specific contingency, the system is considered transiently insecure if the transient stability index (TSI) defined by \ref{TSI} is less than 10\%. In equation \ref{TSI}, $\triangledown\delta_{max}$ is the maximum angular separation between any two rotor angles in degrees. The TSI in Equation \ref{TSI} is based on the TSAT power swing-based algorithm \cite{liu2018accurate}.

\begin{equation}
\label{TSI}
    TSI = \frac{360-\triangledown\delta_{max}}{360+\triangledown\delta_{max}} * 100%
\end{equation}

\item \textbf{Small-signal stability:} Entergy requires a 3\% damping ratio \cite{genc2010decision}. This criterion is applied to the inter-area oscillation modes of generator rotor angles, with a frequency range of 0.25-1.0 Hz and varying amplitudes.
\item  \textbf{Voltage Stability:} A system is considered insecure if any bus voltage deviates from the range of 0.8 pu to 1.1 pu for more than 0.5 seconds \cite{liu2013systematic}.
\item \textbf{Static Security: }The overload index, as calculated in equation \ref{static}\cite{sevilla2015static}, was taken into account.

\begin{equation}
\label{static}
    f_x = \sum_{i=1}^{N_l}wf_i(\frac{S_{mean,i}}{S_{max,i}})^p
\end{equation}

Where $f_x$ represents the overload performance index for the operating point x, $N_l$ denotes the total number of transmission lines, and $S_{mean,i}$ and $S_{max,i}$ indicate the average and maximum apparent power flows of the $i$-th line.

\end{itemize}

\subsubsection{Efficient Database Generation}

While TDS is an accurate and effective method for database generation, it has its own limitations. The computational complexity limits the number of operating conditions it can consider, which hinders its scalability, and thus limits the size of system to which it can be applied for security assessment. As a result, there has been a need to develop an efficient, modular and parallelizable approach for generating databases, that are more easily scalable to larger power systems  \cite{thams2019efficient, venzke2021efficient}.

In light of this limitation researchers have started employing infeasibility certificates to eliminate large portions of the search space of power system operating conditions, if they are practically infeasible. After significantly reducing the search scope, the security criteria is determined (i.e transient stability, voltage stability, etc) where a detailed security boundary is built upon this criteria. Consequently, more samples will be drawn close to a security boundary using historical observations. This technique allows for high parallelization as multiple processes can run concurrently to sample at different security boundaries. This approach was first proposed by \cite{thams2019efficient} and then extended to include small-signal stability in \cite{venzke2021efficient} by assigning the step direction according to the sensitivity of the damping ratio.
%The authors in \cite{venzke2021efficient} conducted remarkable work on creating and publishing an efficient database generator for power system applications in general, and security assessment specifically. They developed a modular and parallelizable approach for generating databases that employs infeasibility certificates to eliminate large portions of the search space (input space of a power system), if they are practically infeasible. Subsequently, they obtain a detailed security boundary description based on the desired security criteria implemented. The approach then creates balanced datasets by sampling near the secure space using historical observations. These aspects make the proposed method fast, efficient, and comprehensive, taking into account historical data features and intelligent sampling techniques to expand the data set as much as possible. 
% pAY ATTENTION TO THE MATHEMATICAL EQUATION AND DIFFERNTIATE BETWEEN THE TYPES OF MULTIPLICATIONS. 

\subsection{Machine Learning Algorithm Training}
\label{ML}
Once the necessary database for training has been constructed, several steps are needed to ensure efficient and accurate training of a machine learning algorithm. These steps encompass data cleaning and pre-processing, feature engineering and extraction, after which the data is ready for training the model, as depicted in Fig. \ref{fig:ML-data-tranining}:

\begin{figure}[h]
    \centering
    \includegraphics[width=0.35\textwidth, height=4.5cm]{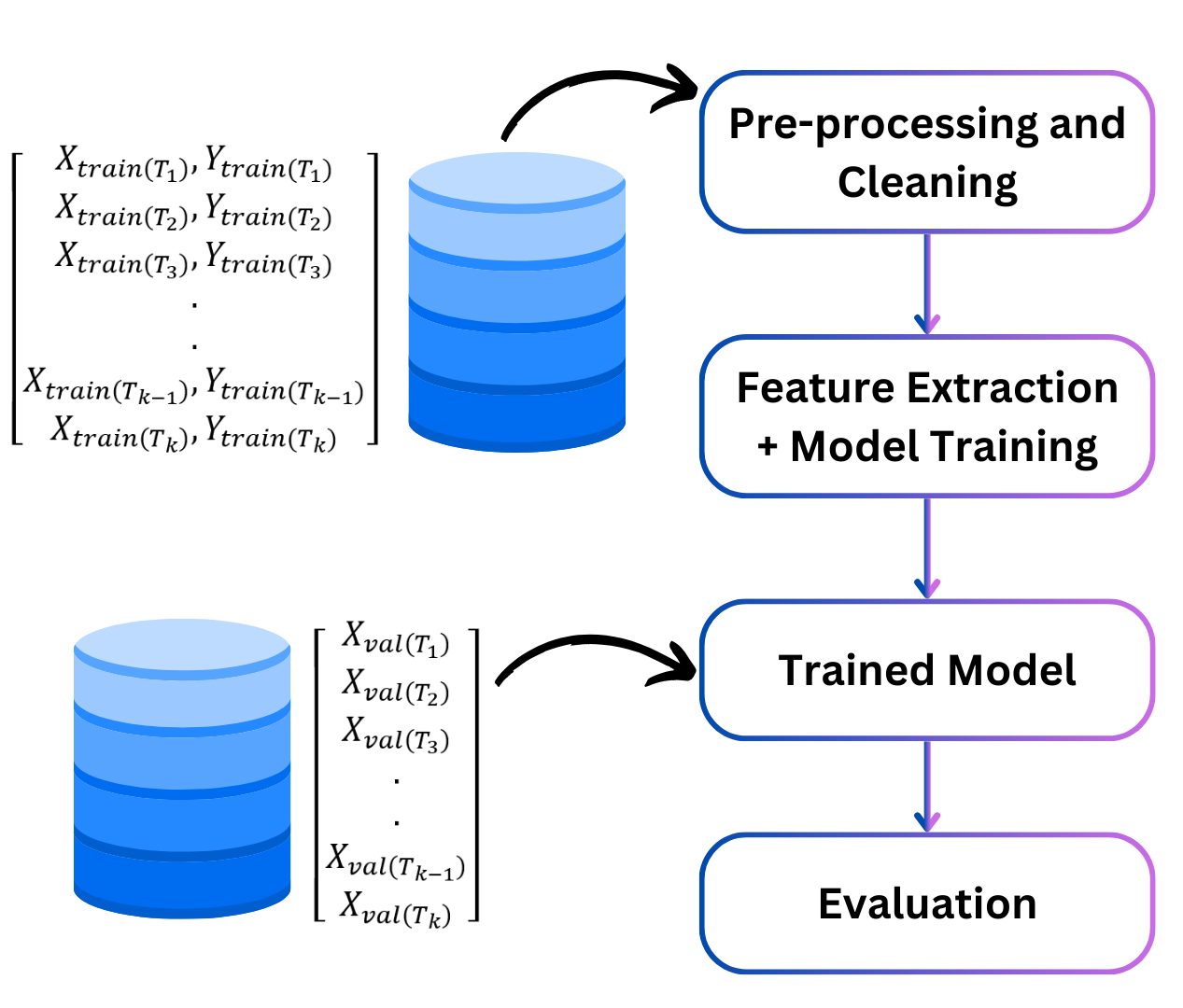}
    \caption{Machine Learning model training stage}
    \label{fig:ML-data-tranining}
\end{figure}

% Pay attention to the mathematical notation in the below paragraph.

In Fig. \ref{fig:ML-data-tranining}, each database represents the generated samples for a single topology $T$, along with their corresponding binary labels (secure/insecure) based on the previously mentioned criteria. Each training $X$ consists of pre-fault operating conditions, including active and reactive power, power flows, phase angles and voltages for each bus. The total number of features is $X \in R^{(C \times n) \times m}$, where C represents the number of contingencies, $n$ denotes the number of operating conditions from one topology, and $m$ is the sum of the number of generators, loads, power flows, voltages, and phase angles in the system. The output label $Y \in R^n$ where $Y$ is a vector in the size of the data representing the labels for each operating condition.

\subsubsection{Proposed Machine Learning Algorithm}

\begin{figure}[h]
\centering
\includegraphics[width=0.45\textwidth]{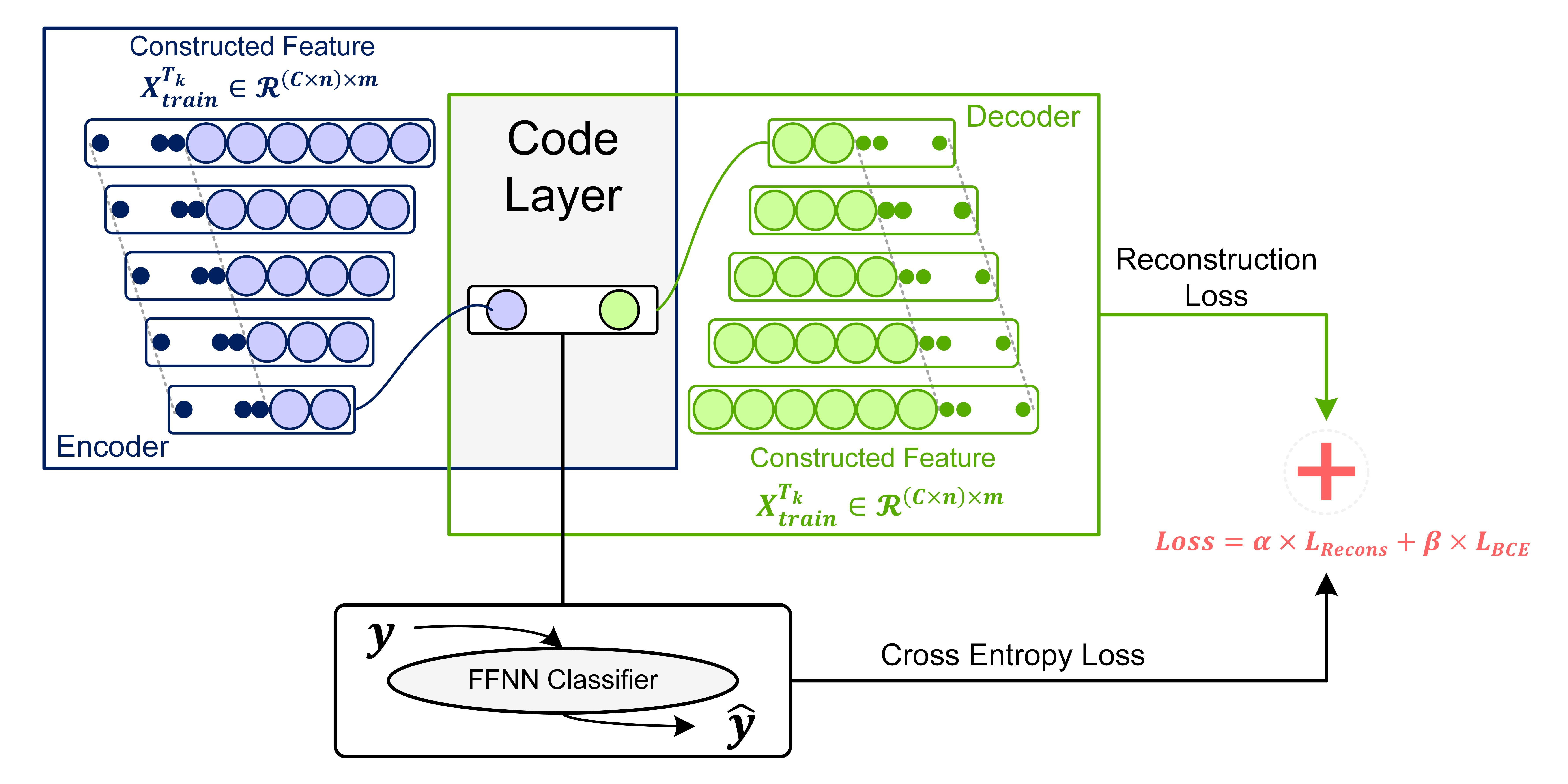}
\caption{Multi-task Semi-supervised Conditional masked Deep Auto-encoder}
\label{fig:mutli-task-learning}
\end{figure}

In this work, we propose a novel technique that integrates the capabilities of conditional masked deep auto-encoders that is trained jointly with the neural network based classifier, named Semi-supervised Multi-Task Learning (SS-MTL), as shown in Fig. \ref{fig:mutli-task-learning}.

This proposed method utilizes a multi-task learning approach, which involves training two tasks concurrently and minimizing a joint loss function across these tasks. The two tasks include feature representation through the conditional deep auto-encoder and the neural network classifier. By utilizing this technique, we create a classification-aware feature representation specifically tailored to our classification objective. In brief, deep auto-encoders convert input features into a dense vector that represents the entire set of input features (power system operating condition). The conditional masking of these auto-encoders improves performance across multiple contingencies. The resulting dense vector representations from the conditional auto-encoders are then fed into a classifier, which is trained to distinguish between secure and insecure power system states. Algorithm 1 demonstrates how the proposed framework works which will be explained in later sections in detail.

\begin{algorithm}[!h]
\caption{SS-MTL Framework}
\begin{algorithmic}[1]
\STATE Define Learning rate, dropout, data size, batch size, Adam optimizer, etc.
\ENSURE Each batch $B$ contains samples only from one topology.
\STATE Initialize all parameters
\textbf{Function}(SS-MTL)($X_{train}^{T_0}$,$Y_{train}^{T_0}$)
\REPEAT
    \STATE
        \STATE Assign 0 to $\beta$ and 1 to $\alpha$
        \REPEAT
            \STATE Mini-batch $M$ optimization 
        \UNTIL{2 Epochs}
\UNTIL{Convergence}
    \REPEAT
        \STATE Mini-batch $M$ optimization 
        \STATE Calculate derivation w.r.t. $\theta$ in (6)
        \STATE Update $\theta$: $\theta = \theta + \lambda \delta \hat{\theta}$
    \UNTIL{$\theta$ has been optimized}
    \STATE \RETURN $f_{\hat{W}}$
\STATE Feed-forward inference through $f_{\hat{W}}^{SSMTL}$ using $X_{test}^{T_k}$, $Y_{test}^{T_k}$

\textbf{Function}(Distance measure)($D$, $th$)
    \STATE Calculate $D$
    \STATE \RETURN Confidence
\IF{New topology}
    \STATE Define Threshold $h$
    \textbf{Function}(Similarity Index)(power system data)
        \STATE Calculate SVS
        \STATE Calculate RMSE
        \IF{RMSE $>$ $h$}
            \STATE Train model again
        \ELSE
            \STATE Keep Model
        \ENDIF
\ENDIF
\end{algorithmic}
\end{algorithm}

\paragraph{Conditional masking auto-encoder}

Auto-encoders are trained with the objective of reconstructing the input data, and are commonly used as a method of feature representation in a lower dimension than the input dimension. It consists of encoder and a decoder networks, where the encoder is responsible for converting the higher dimensional input into a dense lower dimensional vector (called latent representation), whereas the decoder generates an output from the latent representation that attempts to match the input vector.

\textbf{Encoder network}
The encoder network consists of multiple feed forward layers, each applying a nonlinear activation function to a linear transformation of its input:

\begin{equation} h_i = f(W_i \cdot h_{i-1} + b_i) \end{equation}

where $h_{i}$ is the output of layer $i$, $W_i$ and $b_i$ are the weight matrix and bias vector for layer $i$, and $f$ is a nonlinear activation function, such as the ReLU or sigmoid function \cite{lecun2015deep}. The input to the first layer is $h_0=x$ the input data. The latent representation, $z$, is the output of the final encoder layer, $z = h_L$, where $L$ is the number of layers in the encoder network.

\textbf{Decoder network:}
The decoder network is the inverse of the encoder network, where it also consists of multiple layers, each applying a nonlinear activation function to a linear transformation of its input:

\begin{equation} g_i = f(V_i g_{i-1} + c_i) \end{equation}

where $g_i$ is the output of layer $i$, $V_i$ and $c_i$ are the weight matrix and bias vector for layer $i$, and $f$ is a nonlinear activation function. The input to the first layer is $g_0 = z$, the latent representation. The reconstruction, $\hat{x}$, is the output of the final decoder layer, $\hat{x} = g_M$, where $M$ is the number of layers in the decoder network.

The conditional part of the auto-encoder is added in order for the model to be more aware of the different contingencies. Given input data $x$, we first condition the auto-encoder on some auxiliary information $c$, where this information represents the label of the contingency in the training, in the form of a one-hot encoder. The encoder maps the input $x$ and the conditioning information $c$ to a latent representation $z$ as $
z = f_\text{enc}(x, c; \theta_\text{enc})$ where $f_\text{enc}$ is the encoder function and $\theta_\text{enc}$ represents its parameters. The decoder then maps the latent representation $z$ and the conditioning information $c$ back to the reconstructed input $\hat{x}$ as $\hat{x} = f_\text{dec}(z, c; \theta_\text{dec})$ where $f_\text{dec}$ is the decoder function and $\theta_\text{dec}$ represents its parameters.

\begin{equation}
L_{recon} = ||x - D(E(x, c), c)||^2
\end{equation}
where $x$ is the input data and $C$ represents the one-hot encoded representation of the contingencies, whereas the encoder $E$ maps the input to a latent representation $z$, and the decoder $D$ reconstructs the input from the latent representation.

\paragraph{Supervised classifier} 

The neural network classifier is trained using the latent representation $z$ obtained from the encoder. The classifier aims to minimize the classification loss $L_{class}$:

\begin{equation}
L_{class} = - \sum_{i=1}^{N} y_i \log C(E(x_i, c_i))
\end{equation}

Here, $N$ represents the number of labeled samples, $y_i$ is the true label for the $i$-th sample, and $c_i$ is the conditioning information for the $i$-th sample.

To train these two tasks simultaneously on the combined loss function, which is a combination of the reconstruction loss $L_{recon}$ and the classification loss $L_{class}$, hyper-parameters $\alpha$ and $\beta$ are used to balance the importance of the two losses:

\begin{equation}
\label{accumulative}
L = \alpha L_{recon} + \beta L_{class}
\end{equation}

The goal of the semi-supervised architecture is to minimize the combined loss $L$ by optimizing the parameters of the encoder, decoder, and classifier. This results in a model that leverages both the unsupervised reconstruction task and the supervised classification task to improve the overall performance. The training process for this method proceeds as follows:
\textit{(i)} Auto-encoder warmup, where initially the auto-encoder is trained exclusively by setting the $\beta$ in \ref{accumulative} to 0 and $\alpha$ to 1. This step is introduced to prevent the classifier network from training on representations when they are closer to random values than representative weights.
\textit{(ii)} The values of $\alpha$ and $\beta$ are adjusted to 0.5, which prompts the optimizer to update both the classifier and auto-encoder weights.
\textit{(iii)} In the final stage, the values of $\alpha$ and $\beta$ are set to 0.75 and 0.25, respectively, allowing for the fine-tuning of the classifier for secure/insecure predictions.

\subsection{Topological Similarity Index Model}
\label{similarity-sec}
Topological changes in a power system can arise due to various reasons, such as planned maintenance, equipment failures, or the integration of renewable energy sources. These changes can significantly affect the performance of a machine learning-based power security assessment too, as it might alter the behaviour of such systems. Therefore, we propose adding an additional layer that detects topological similarity and evaluates the need to re-train the algorithm on the newly introduced topology.

Multiple indices of topological similarity have been proposed and evaluated in the literature \cite{bush2021topological, cao2020study} for power systems applications, with the Singular Value Sequence (SVS) being recognized as the most effective in terms of accuracy and efficiency \cite{cao2020study}. SVS is calculated using singular value decomposition (SVD) \cite{cao2020study, lecun2015deep}. %Singular Value Decomposition (SVD) is a technique used to factorize a matrix into three constituent matrices. Where given a matrix $A$ with size $(m \times n)$, SVD decomposes A into three matrices: $U$ $(m \times m)$, $\sum$ $(m \times n)$, and $V^*$ $(n \times n)$, where $U$ and $V^*$ are orthogonal matrices and $\sum$ is a diagonal matrix containing singular values \cite{cao2020study}.
%
% Pay attention to the correct math
%
%    \begin{equation}
%    \label{bad-data}
%A = U \cdot \Sigma \cdot V^{*}
%\end{equation}
%
%The singular values are the square roots of the non-negative eigenvalues of $A$ or $A\cdot A^T$. They are arranged in descending order along the diagonal of $\sum$, i.e., $\sigma_1 \ge \sigma_2 \ge ... \ge \sigma_n \geq 0$. 
The singular value sequence (SVS) is the ordered set of singular values. In the context of power systems, SVS offer insights about the power system characteristics that have been utilized in previous applications such as fault detection and diagnosis, and stability analysis.

For power system topological similarity between two different topologies, we employ the Root Mean Square Error (RMSE) between the SVS of two different topologies to quantify their similarity. The lower the RMSE value, the more similar the topologies. Let SVS $S_1 = \{ \sigma_{11}, \sigma_{12}, ..., \sigma_{1n}\}$ and $S_2 = \{ \sigma_{21}, \sigma_{22}, ..., \sigma_{2n} \}$ be the singular value sequences of the two different power system topologies. The RMSE between $S_1$ and $S_2$ can be calculated as follows:

% Very commonly use statement should be backed by a citation.

\begin{equation}
\label{rmse}
\text{RMSE}(S_1, S_2) = \sqrt{\frac{1}{n} \sum_{i=1}^{n} (\sigma_{1i} - \sigma_{2i})^2}
\end{equation}
where $\sigma$ are the singular values obtained from the SVD.
\subsection{Distance Measure}
\label{distance}
Finally, after the predictions are generated through the classifier, we employ the Mahalanobis distance \cite{mili1993robust} to estimate the uncertainty in the prediction, in order to provide the operator with insights into the prediction, as well as establish confidence in their reliability. The Mahalanobis distance is a metric used to measure the distance between two data points in a multivariate space while taking into account the correlations between variables. It is particularly useful for measuring distances between data points when the variables have different scales or units. The Mahalanobis distance has been utilized to measure the distance between two operating conditions in a power system by considering the correlations between power system variables and the different scales at which they operate \cite{pagnier2021physics}. The Mahalanobis distance between two data points $x$ and $y$ is defined as \cite{jordan2015machine}:

\begin{equation}
D_{M}(x, y) = \sqrt{(x - y)^{T} \cdot S^{-1} \cdot (x - y)}
\end{equation}

where $S$ is the covariance matrix of the data, and $S^{-1}$ is its inverse. The covariance matrix captures the correlations between the different variables in the data, and the inverse of this matrix is used to normalize the distance calculation.

In the context of power systems, operating conditions can be represented by a set of variables such as voltage magnitudes, angles, active and reactive power injections, and power flows. When measuring the distance between two operating conditions, the Mahalanobis distance takes into account the correlations between these variables and the different scales at which they operate, providing a more meaningful measure of distance than the Euclidean distance \cite{mili1993robust}.

To provide confidence in the predictions of the network state, the above-mentioned distance is used to measure the distance between the current operating point (where the model generated the label) and all of the operating points in the dabase on which the model was trained. This assumes that it is more likely that the trained model will generate more accurate predictions if the current operating point is close to the trained operating conditions. Nevertheless, comparing the current operating point with all the operating points in the database may not result in a significant difference when all the distances are averaged. For this reason, we calculate the average of the Mahalanobis distances between the current operating condition and its 50 nearest points. This average is then used to determine the confidence level associated with the current condition.
\subsection{Model Evaluation:}
\label{evaluation}
Power system security assessment is a safety critical application where the cost of missed alarms out-weighs the cost of false ones. Therefore we need to establish a metric  which penalizes missing alarms. Resilience against imperfect data is also an important aspect of a strong framework, hence we will evaluate the algorithm's resilience to better understand how it would perform under realistic operating conditions with measurement uncertainty, communication delays, and missing data.

\subsubsection{F-beta Score:}

The F-beta score is a measure of a classification model's accuracy, which considers both precision ($\phi$) and recall ($\rho$). It is a weighted harmonic mean of precision and recall, with the beta parameter determining the importance of recall relative to precision. The F-beta score is given by \cite{lecun2015deep}:

\begin{equation}
F_\beta = (1 + \beta^2) \frac{\mathrm{\phi} \cdot \mathrm{\rho}}{(\beta^2 \cdot \mathrm{\phi}) + \mathrm{\rho}}
\end{equation}

The selection of $\beta$ was chosen to be 2 since, in the context of power system security assessment it emphasizes penalizing missed alarms greater than false alarms. This is because missed alarms could result in significant reliability repercussions, i.e. potential blackouts, while false alarms, although not harmless, may merely cause operators to take unnecessary preventive actions.

\subsubsection{Bad Data Resilience}

In practical scenarios, it is frequent to encounter inaccurate or missing data. Thus, it is crucial to assess the ability of algorithms to manage poor data, whether it results from faulty measurements, or external malicious attacks \cite{abedi2019review}. Bad data in general exploits the vulnerabilities of machine learning models in terms of their resilience, particularly deep neural networks \cite{narodytska2017simple}, which might cause them to produce incorrect outputs.

In order to simulate such conditions, we will introduce perturbations to the input data, which are designed to deceive the model while appearing virtually identical to the original input. A common approach to generating adversarial examples to simulate bad data is the Fast Gradient Sign Method (FGSM)\cite{ren2021vulnerability, zheng2021vulnerability}. Given a model with loss function $J(\theta, x, y)$, where $\theta$ represents the model's parameters, $x$ is the input data, and $y$ is the true label, the bad data example $x_{adv}$ is generated as follows \cite{abdi2010principal}:

\begin{equation}
\label{bad-data}
x_{\text{adv}} = x + \epsilon \cdot \text{sign}(\nabla_x J(\theta, x, y))
\end{equation}

Here, $\epsilon$ is a small constant that controls the magnitude of the perturbation, and $\nabla x J(\theta, x, y)$ is the gradient of the loss function with respect to the input data x. The sign function returns the element-wise sign of the gradient, ensuring that the perturbation is small yet effective. Opting for minor perturbations over significant ones is due to the fact that large disturbances are easily identified by both the algorithm and the model, making it less likely for them to mislead the model. In this approach, a neural network was trained using a portion of the original training data to produce undesirable data examples using equation \ref{bad-data}. It is important to note that this equation is frequently utilized to create adversarial examples \cite{pan2021improving}, which rank among the most prevalent instances of flawed data or efforts to trick the model.

\section{Implementation of SS-MTL and Testing Methodology}

\subsection{Implementation}
%% Write all about how you implemented your method, in what environment, and detailed description of how ML end worked, training, online running, paramter selection, etc.
\subsubsection{Database Construction}

% All of the notations should be represented as single letters instead of statements

The initial step is constructing the database, encompassing (i) pre-fault operating conditions (OCs) data and (ii) corresponding post-fault labels that identify whether the system is safe or not. The pre-fault OCs encompass active and reactive power (either generation $G^{or}_{ac} \in R^{n\times g}$, $G^{or}_{re} \in R^{n\times g}$ or load $L^{or}_{ac} \in R^{n\times l}$, $L^{or}_{re} \in R^{n\times l}$, power flows $F^{or}_{ac} \in R^{n\times f}$, $F^{or}_{re} \in R^{n\times f}$, voltages $V_{\text{or}} \in R^{n\times v}$, and phase angles $\Theta_{\text{or}} \in R^{n\times \theta}$ for each bus. Combined, these simulations form the $m$ dimensional original training features $X_{\text{or}} \in R^{n\times m}$, where $n$ represents the size of the entire dataset for one topology and $m = 2 \times (g + l + f) + v + \theta$. The corresponding post-fault labels are denoted as $Y_{\text{or}} \in R^{n}$, where each element is denoted as $y_i$, where $y_i$ is assigned 1 when secure and 0 otherwise, and is computed based on the aforementioned indices provided in \ref{generation}. We use $X^{T0}_{\text{train}}$, $Y^{T0}_{\text{train}}$, $X^{T0}_{\text{test}}$, and $Y^{T0}_{\text{test}}$ to denote the data from the first topology $T0$.

\subsection{Testing Methodology}

\subsubsection{Case Study: 68-bus Power System}

The IEEE 68-bus power system is a well established benchmark that contains 16-machines across 5 regions which is a reduced equivalent of the interconnected New England test system (NETS) and New York power system (NYPS) \cite{pal2006robust, anagnostou2015impact}.
For the sake of simplicity, we assume that PMU devices are installed and provide real-time measurements. Therefore, at each bus voltage magnitudes, phase angles, active and reactive power are available. On the other hand, the power flow data is calculated by the solver while generating the data. Following \cite{zhang2021confidence}, for generating and sampling observations from a set of predefined operating conditions, we drew the active load power from a multivariate Gaussian distribution. The active load power is scaled within +-50\% if the nominal values, while the reactive power was scaled by the assumption of having a constant impedance for the buses. The power factor was restricted to $[0.95,1]$. 

The contingencies considered are selected following the rules stated in \cite{pal2006robust}, whereas the location of the faults were selected to be close to the generators, while the clearing of the fault is coupled with line tripping. Examples of the considered contingencies can be shown in Table III in \cite{anagnostou2015impact}. To generate operating conditions from variety of topologies from IEEE 68-bus system \cite{pal2006robust}, we have followed the works in \cite{zhang2021confidence, arteaga2019deep, zhang2019deep} that generated 44 different topologies. Examples of different topologies are the double line connection between bus NO.27 and NO.53 which suggests that any disconnection between these two buses would likely have a minor impact on the rest of the network. Bus NO.17 carries the highest load in the system, indicating that a disconnection at this bus could result in significant changes to the power flow patterns \cite{zhang2021confidence}.

\subsubsection{Testing Procedure}

% Time-Domain-simulation method to for sample generation was implemented using Python code with muliple packages; Pandapower, Pyomo, PyPSA and Numpy. Both of the above data generation techniques were executed on a AWS instance that contains 84 CPUs and 16 GB of RAM. Machine learning based algorithms were implemented using Pytorch package and the training of these algorithms was carried on an instance with 16 CPUs for decision trees and support vector machines, while other deep learning based algorithms were trained using Nvidia T4 GPU. 

The time-domain simulation method for sample generation was implemented in Matlab and executed on an AWS instance with 84 CPUs and 16 GB of RAM. Machine learning algorithms were developed in Python \cite{githubGitHubMuhizatarMachinelearningforpowersystemsecurityassessment}. Training was conducted on an instance with 16 CPUs for decision trees and support vector machines, while deep learning algorithms utilized an Nvidia T4 GPU.

To ensure a thorough and fair assessment of machine learning algorithms for power system security, the training and testing processes must be comprehensive to include all possible scenarios such as unseen topologies. Some of the generated topologies were included in both the training and testing databases, allowing for an evaluation of the algorithms’ performance on familiar topologies. Additionally, to assess performance on new, unseen topologies, certain topologies were exclusively reserved for testing. Furthermore, during the training phase, each batch provided to the model contained observations from only one topology.

\section{Numerical Results and Discussion}

% This is where we need to switch the writing to be more power-system oriented,

% This section needs to be written differently.

% Include a graph that describes the IEEE-68 bus systems

% Mention how the 22 topologies were generated.

% Write in detail all the IEEE
The algorithms were assessed across 22 distinct topologies for the IEEE 68 bus system on 4000 overall test samples. It is worth noting that two topologies were intentionally selected for all the aforementioned systems to significantly deviate from the default topology. One of these topologies was included in both the training and testing operating conditions, while the other was only incorporated in the testing phase to evaluate the effectiveness of the similarity model. The error bars in the results figure illustrates the variance of the scores across the 22 different topologies.

\begin{figure}[!h]
  \vspace{-1em}
    \centering
    \includegraphics[width=0.45\textwidth, height=7.5cm]{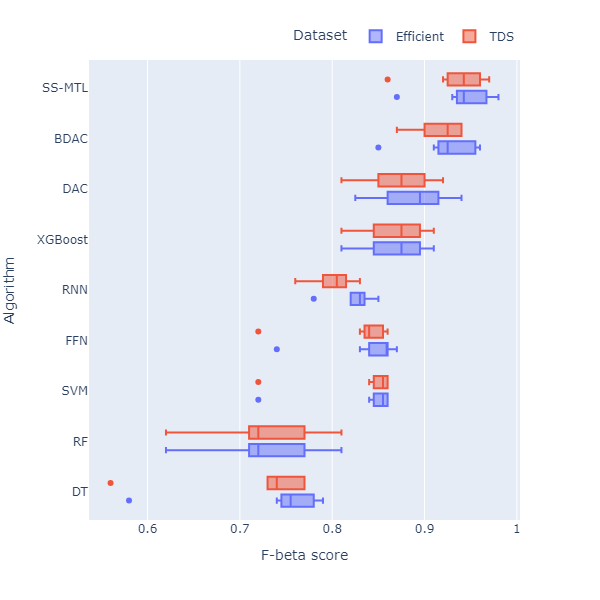}
    \vspace{-2em}
    \caption{Performance of the algorithms on the IEEE 68 bus system}
    \label{fig:performance-basic}
     \vspace{-2em}   
\end{figure}

Fig. \ref{fig:performance-basic} demonstrates that the proposed algorithm outperforms all other algorithms in terms of F2 score on both database generation techniques. The average F2 score on all topologies for the proposed algorithm is around 0.95, while the BDAC algorithm proposed in \cite{zhang2021confidence} has a 0.925 average F2 score. Machine learning based classifiers such as DT and RF had average F2 scores of 0.74 and 0.72, respectively.  SVM had the best performance of this category of algorithms with a 0.85 F2 score. Deep learning based methods without auto-encoders outperformed DT and RF and had similar performance to SVM methods with 0.84 and 0.81 for FFNN and Recurrent Neural Networks (RNNs) respectively.

From Fig. \ref{fig:performance-basic} several observations can be made. First of all, the performance of neural networks and their variants surpasses DT, RF, and SVM, presenting a distinct trade-off between accuracy and interpretability. Techniques that draw a decision boundary, i.e. DT, RF, and SVM, are significantly more affected by the unseen topology (lowest score for each bar in the figure). Algorithms that employ auto-encoders (deep , convolutional, variational auto-encoders, etc)  for feature extraction demonstrate significantly better performance compared to classifiers implemented on raw data. On the database generation side, the efficient database generation technique produced improved results across almost all algorithms, which may be due to the larger database or the reduced complexity in the data, as transient stability is not considered in the criteria. However, the performance difference between the algorithms follows the same pattern for both database generation techniques, which underscores the importance of the database generation method in enriching the number and quality of samples. Finally, the ensemble method (XGBoost) displays exceptional performance on both dataset generation techniques, despite being trained on raw data. This suggests that future work exploring this technique combined with improved feature selection could potentially achieve a balance between interpretability and accuracy.

\subsection{Reliability and Vulnerability Assessment}

\begin{figure}[!h]
  \vspace{-1em}
    \centering
    \includegraphics[width=0.45\textwidth, height=7.5cm]{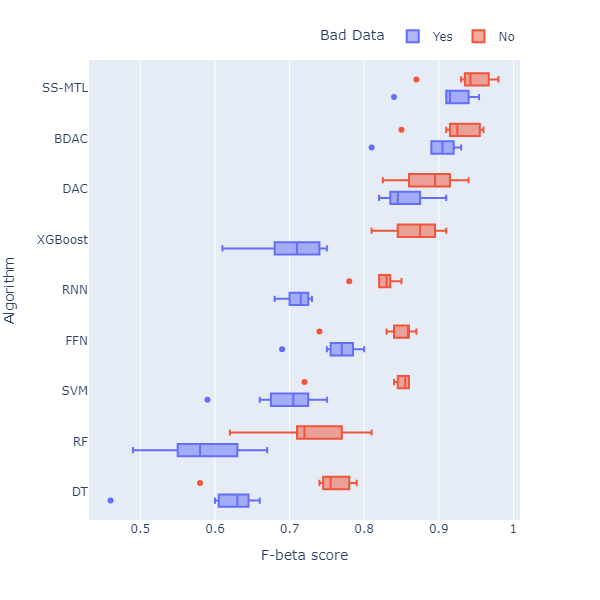}
    \vspace{-2em}
    \caption{Performance of the algorithms on the IEEE 68 bus system, with and without bad data}
    \label{fig:performance-bad-data}
    %\vspace{-2em}    
\end{figure}

To assess the vulnerability of the trained algorithms to bad data, we evaluated the algorithms' performance with and without the injection of bad data in the test-set. Fig. \ref{fig:performance-bad-data} displays the performance of the various algorithms on the IEEE 68 bus system for both database generation algorithms. Small perturbations in the input significantly impacted the performance of classifiers such as DT, RF, SVM, and XGBoost. This is expected since they establish hard decision boundaries, thus making them susceptible to minor perturbations in the input. Feed-forward Neural Networks and Recurrent Neural Networks displayed better robustness compared to other classifiers. However, autoencoder-based techniques demonstrated superior resilience in the presence of bad data, aligning with the literature \cite{ren2021vulnerability, chen2018machine}.  Methods that draw a hard boundary across the decision boundary (i.e. DT, RF, SVM, etc.) were more affected by bad data as their F2 Score tended to drop by 0.2 points between the cases, whereas neural network based methods such as RNN and FFNN showed some drop in F2 score performance (0.14 and 0.17, respectively), but less relative to the hard boundary methods. Finally, auto-encoder methods \cite{zhang2021confidence, zhang2021deep}, which are known for their resiliency \cite{wang2016auto}, showed a drop of only 0.03 in F2-score between the cases.

The performance comparison of the implemented algorithms shows that DT and RF provide moderate accuracy and varying levels of interpretability, robustness, and scalability, with low training time and parameter tuning complexity. SVM have moderate accuracy and scalability but low interpretability, with higher training time demands. XGBoost delivers high accuracy and scalability, with moderate robustness and training time, and requires moderate parameter tuning complexity. FFNN and RNN also offer high accuracy and scalability, with moderate robustness and similar training time, though they require high parameter tuning complexity. DAC and CBDAC, along with the proposed SS-MTL (which achieves the highest accuracy), stand out with very high accuracy, robustness, and scalability but are complex to tune and demand more data and time for training. These models are highly effective but computationally intensive.

\subsection{Similarity Index Integration}

As previously mentinoed, we included 22 different topologies in the test, 21 included in the training and one that was not included in the training. In both Figs. \ref{fig:performance-basic} and \ref{fig:performance-bad-data} it is evident that for most of the algorithms, there is one outlier value with performance vastly worse than the others. In these cases it is the out-of-training topology, indicating a need to update the model in case of a significant topology change. In this study, we have added topological awareness to machine learning frameworks by incorporating a similarity index model (\ref{similarity-sec}. This model, which depends on calculating the SVS between the newly introduced topology and those in the training set, helps determine whether the machine learning algorithm's weightings need to be updated to accommodate the new topology. Therefore, it is necessary to establish a similarity threshold, determined empirically by analyzing the relationship between the Root Mean Square (RMS) difference (equation \ref{rmse}) and its effect on the F-beta score. To compute the mean RMSE for any new topology, we average the RMSE values between the new topology and all existing topologies.
%To demonstrate the process of estimating the threshold, we present the singular value sequence for the IEEE 14 bus system in the table below, along with the corresponding decrease in the F-beta score when compared to the default topology.

% In the example shown in the table below, we removed one topology from the training data, and computed the RMSE between their SVS. For the F-beta score, we first tested the model on the test sets of the trained topologies, averaged the score, and then tested the model on the unseen topology to measure the decline in the F-beta score of the other topologies.

%\begin{table}[h]
%\centering
%\caption{Mean RMSE and its correlation with the F-beta score}
%\begin{small}
%\begin{tabular}{|c|c|}
%\hline
%\textbf{Mean RMSE} & \textbf{F-beta score difference} \\ %\hline
%         2.3          &     $  4.5\times  10^{-2} $        %  \\ \hline
%        2.7        &     $  3.4 \times  10^{-2} $          % \\ \hline
%          8.4         &     $    13.2 \times  10^{-2} $           \\ \hline
%          3.7         &     $    5.1  \times  10^{-2} $    %        \\ \hline
%          7.6         &     $    9.4  \times  10^{-2} $    %      \\ \hline
%\end{tabular}
%\end{small}
%\end{table}

Based on empirical results, the similarity threshold for the IEEE 68 bus system was determined to be 16; when higher, the model weights need to be updated, when lower it is safe to continue with the same model.
%Based on Table II, the threshold for the similarity model was set at 4 for the IEEE 14 bus system. Comparable experiments were conducted for the IEEE 39 bus and IEEE 68 bus systems, where the thresholds were set at 10 and 16, respectively.

\begin{figure}[!h]
    \vspace{-1em}
    \centering
    \includegraphics[width=0.45\textwidth, height=7.5cm]{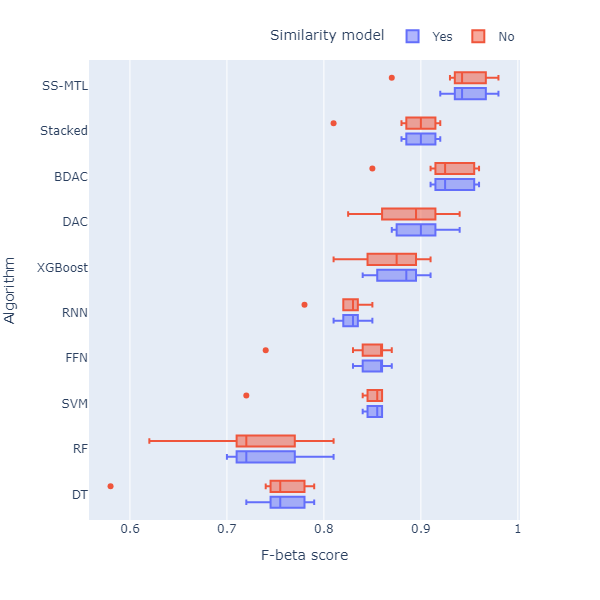}
    \vspace{-2em}
    \caption{Performance of the algorithms with the similarity model}
    \label{fig:performance-smiliar-model}  
\end{figure}

As seen in Fig. \ref{fig:performance-smiliar-model}, incorporating the topological similarity model, improved the model's awareness of unseen topologies and removed the topology with the lowest score from the training data for the updated model. It should be noted, however, that the scores for other topologies changed only slightly, remaining within a similar range. The consistent behavior demonstrated by all algorithms attests to the efficacy of the similarity model. Finally, it is important to note that these experiments were conducted exclusively with TDS.

\subsection{Speed Comparison}

Since the motivation behind using machine learning for dynamic security assessment is driven by the reduced computation speed that would allow real-time predictions, it is essential to compare the speed of prediction between TDS (the most accurate method for dynamic security assessment) and machine learning techniques. Table II shows the time required for TDS and SS-MTL for different systems.

\begin{table}[h]

\centering
\begin{small}
\caption{Speed Comparison between TDS and SS-MTL}
\begin{tabular}{|c|c|c|}
\hline
\multicolumn{1}{|c|}{\begin{tabular}[c]{@{}c@{}}System\end{tabular}} & \multicolumn{1}{|c|}{\begin{tabular}[c]{@{}c@{}}Time (ms) for TDS\end{tabular}} & \multicolumn{1}{|c|}{\begin{tabular}[c]{@{}c@{}}Time (ms) for SS-MTL\end{tabular}} \\ \hline
IEEE 14 Bus                                                             &         2450        &          45                                       \\ \hline
IEEE 39 Bus                                                             &         8640        &          67                                          \\ \hline
IEEE 68 Bus                                                             &        19560         &           87                                         \\ \hline
Nesta 162 Bus                                                           &         25780        &             122                                       \\ \hline
\end{tabular}
\end{small}
\end{table}

As can be seen, SS-MTL provides approximately 200 times speed up in security assessment. It is important to note that these TDS results are per contingency, whereas machine learning results are for 22 contingencies per operating condition. It is worth noting that the time complexity of TDS with respect to the number of buses is $O(N^3)$ (could vary based on the solver used) where N is the number of buses in the system, suggesting challenges in scaling to interconnection-scale systems. The complexity of a trained neural network is proportional to number of layers and neurons, not the size of the power system, and therefore it will increase very little when scaled to larger power systems.
\section{Conclusion}

%This paper addresses the problem of dynamic security assessment in power system by employing machine learning-based techniques. It developed a Semi-Supervised Multi-Task Learning (SS-MTL) framework in order to enhance the performance, to improve topological awareness, and enhance resilience against bad data. Evidenced by the results presented, the proposed algorithm surpassed existing state-of-the-art machine learning methods in power system security assessment, especially improving the performance on missed alarms. The developments of this paper can be employed as a preliminary screening phase for power system security assessment, offering the operators with insights into the state of health of the grid given credible contingencies. Future research could further emphasize  minimizing the occurrence of missed alarms. Additionally, evaluating the algorithm's efficacy on larger systems and authentic historical data will be of paramount importance to move toward industrial adoption.

This paper explores dynamic security assessment in power systems using machine learning techniques. It introduces a Semi-Supervised Multi-Task Learning (SS-MTL) framework to improve performance, topological awareness, and resilience to bad data. Results show the proposed algorithm outperforms existing methods, particularly in reducing missed alarms. This development serves as an initial screening tool for assessing power system security, providing operators insights into grid state under credible contingencies. Future work will focus on reducing missed alarms and testing the algorithm's effectiveness on larger systems to encourage industrial adoption.

\bibliographystyle{IEEEtran}
\bibliography{Muhy_Lib}

% that's all folks
\end{document}